\documentclass[11pt,a4paper]{article}
\usepackage[hyperref]{acl2020}
\usepackage{times}
\usepackage{latexsym}

\usepackage{microtype}

\aclfinalcopy 

\usepackage{tabularx}
\usepackage{multirow}
\usepackage{graphicx}
\usepackage{comment}

\title{Where is the context? -- A critique of recent dialogue datasets}

\author{Johannes E.\ M.\ Mosig \\
  Rasa Technologies GmbH \\
  \texttt{j.mosig@rasa.com} \\\And
  Vladimir Vlasov \\
  Rasa Technologies GmbH \\
  \texttt{vladimir@rasa.com} \\\And
  Alan Nichol \\
  Rasa Technologies GmbH \\
  \texttt{alan@rasa.com} \\}

\date{}

\begin{document}
\maketitle
\begin{abstract}
Recent dialogue datasets like MultiWOZ 2.1 and Taskmaster-1 constitute some of the most challenging tasks for present-day dialogue models and, therefore, are widely used for system evaluation.
We identify several issues with the above-mentioned datasets, such as history independence, strong knowledge base dependence, and ambiguous system responses.
Finally, we outline key desiderata for future datasets that we believe would be more suitable for the construction of conversational artificial intelligence.
\end{abstract}

\section{Introduction}

The recent dialogue datasets MultiWOZ \cite{budzianowskiMultiWOZLargeScaleMultiDomain2018, ericMultiWOZMultiDomainDialogue2019} and Taskmaster-1~\cite{byrneTaskmaster1RealisticDiverse2019} facilitate the construction of task-oriented machine learning dialogue systems, as exhaustively reviewed by \cite{gaoNeuralApproachesConversational2019}, where a user wants to know or do something, while the system has to understand the user's utterance and reply appropriately.
Learning to model these datasets is presently among the most challenging tasks for state-of-the-art dialogue systems, as they cover multiple task domains and contain dialogues in which users change their goal during the conversation.
Nevertheless, both MultiWOZ and Taskmaster-1 suffer from a number of issues, which we discuss in the present paper.

The issues we discuss concern the task of predicting the next action of the system, given the dialogue history.
Here, we define a system action as a combination of task domain (restaurant, hotel, etc.), dialogue act type (inform, recommend, etc.), and slots filled by the system, following \citet{zhangTaskOrientedDialogSystems2019}.

As a first issue we find that, when comparing multiple dialogues, different actions follow from the same dialogue history and thus a system trained on these dialogues cannot learn deterministic behaviour.
Such ambiguous actions can only be handled by systems that are designed to model a distribution of system actions, such as in the recent work of \cite{zhangTaskOrientedDialogSystems2019}, who also acknowledged this issue as a ``one-to-many'' property of human dialogue in general.
Presently, however, most supervised systems only predict a single best response \cite{mehriStructuredFusionNetworks2019,chenSemanticallyConditionedDialog2019,madottoMem2SeqEffectivelyIncorporating2018}.

As a second issue, we observe that dialogue models do not seem to benefit from knowing the dialogue history beyond the last user input and its preceding system action, which indicates an unnatural simplicity of the dialogues.

Throughout this article we follow \cite{mrksicNeuralBeliefTracker2016, wen_network-based_2017} and decouple the training data from the knowledge base, as the system's choice of action should primarily depend on the number of items (hotels, pizza toppings, etc.) that satisfy the user's criteria, not their specific names.

In this article, we do not seek to improve on the performance of state of the art models on the datasets we consider, nor do we compare our models to them. 
Instead, we use different models as a tool to identify problems within these datasets.

We first discuss the MultiWOZ dataset in \S\ref{sec:multiwoz}, and then briefly the Taskmaster-1 dataset in \S\ref{sec:taskmaster}, before we conclude with \S\ref{sec:conclusions} where we summarize our findings and outline ways to circumvent the identified problems.
The program code used in this paper is available under \url{https://github.com/RasaHQ/multiwoz-paper}.

\section{MultiWOZ}
\label{sec:multiwoz}

The MultiWOZ dataset~\cite{budzianowskiMultiWOZLargeScaleMultiDomain2018}, contains over 10k task-oriented conversations on hotels, restaurants, taxi and train bookings, attractions, hospitals, and police stations in the city of Cambridge.
Many of the dialogues cross several of these domains and, on average, span about 14 turns per dialogue.
Most MultiWOZ dialogues come with annotations for the system's action and gold belief state (the user's goal and slot values), which sets it apart from other multi-domain dialogue datasets such as MetaLWOz \cite{schulzMetaLWOzDatasetMultiDomain2019}.

The MultiWOZ dialogues are collected with a Wizard-of-Oz setup~\cite{kelleyIterativeDesignMethodology1984}, where the system's role is taken by a human, ensuring that the system's output utterances are formulated naturally.
In the MultiWOZ setup the wizard \textit{chooses} a response, which is distinct from the paraphrasing approach of \cite{rastogiScalableMultidomainConversational2019}, where the dialogue acts are fixed by a schema, and crowd workers paraphrase the dialogue acts.

All dialogues were collected via Amazon Mechanical Turk \cite{crowston_amazon_2012}.
Turkers in the user's role were asked to want to achieve a certain goal, e.g.\ find a hotel in the city center with free WiFi and then book it (goals were revealed to the turker over time), and turkers in the system's role (wizards) were asked to respond appropriately and check for the availability of requested hotel rooms, etc., via a specially designed user interface.
The conversations were then annotated with action labels and gold belief states of the wizards.
Here we consider the revised version, MultiWOZ 2.1, in which many labeling errors have been corrected manually~\cite{ericMultiWOZMultiDomainDialogue2019}.
Throughout this Section, we split MultiWOZ into a training and a test set at a 80/20 ratio. 


\subsection{Ambiguous system actions}
\label{sec:multiwoz:ambiguous}

In this Section we demonstrate that, when comparing multiple dialogues, we find that different actions follow from the same dialogue history and thus a system trained on these cannot learn deterministic behaviour.
To this end, we train a ``memorization model'' that simply memorizes sequences of events in the training data.
The events are system actions on the one hand, and the user's dialogue acts, represented as tuples of user intents \cite{budzianowskiMultiWOZLargeScaleMultiDomain2018} as well as slots that the user wants to fill, on the other hand.
The number of events that are taken into account for prediction is limited by \verb+max_history+, which we set to 10.
If the memorization model cannot achieve an F1 score of 1.0 on the training set, then the system's actions are ambiguous.

The input and output spaces of the memorization model are the user input and the possible system actions, similarly to the POMDP-based methods \citep{young2013pomdp} and in contrast with end-to-end models. 
We call dialogue models that operate on these simplified input and output spaces ``modular models'', following \cite{wen_network-based_2017}. 

While the system actions are already part of the MultiWOZ annotation, the user intents are not.
Thus, we follow \citet{vlasovDialogueTransformers2019} and define two intents, \verb+inform+ and \verb+bye+.
A user input has the intent \verb+inform+, unless it is the last intent of the dialogue and the user did not provide any slots. 
In this case we assume (based on reading a sample of dialogues) it is a farewell and thus assign the intent \verb+bye+.

We infer the slots provided by the user from changes in the system's belief state after a user input.
For example, a user intent/slot tuple for the phrase ``I'd like to stay at a 4-star hotel'' is thus \verb+inform{"hotel_stars":+ \verb+"specific"}+.
Note, that we do not store the actual number of stars in the example above, but the generic tags \verb+specific+ or \verb+do-not-care+, since the particular star rating of the hotel should not matter for next-action prediction.

In addition, we slightly augment the action labels of the MultiWOZ dataset.
Specifically, we infer the domain (hotel, restaurant, etc.) of the system's \verb+Booking-Book+ action from the last mentioned domain and add this information to the label, resulting, e.g., in \verb+Hotel-Booking-Book+.
A typical dialogue, as seen by a modular model, may now look like the first example presented in Appendix~\ref{apx:example}.

\begin{table*}[t]
    \small
    \centering
    \begin{tabular}{|l|c|c|cc|cc|cc|}
    \hline
                &   & Memorization & \multicolumn{2}{c|}{Modular LSTM} & \multicolumn{2}{c|}{Modular TED} & \multicolumn{2}{c|}{End-to-end TED} \\
                &   & Training    & Training        & Testing  & Training        & Testing       & Training         & Testing         \\
    \hline
    Initial & 10   & 0.84    & 0.51 & 0.51 & 0.50            & 0.49          & 0.22             & 0.21            \\
            & 2    & & 0.47 & 0.47 & 0.47 & 0.45 & 0.22 & 0.21 \\
    Simplified  & 10   & 1.00    & 0.95 & 0.92 & 0.95            & 0.89          & 0.69             & 0.66            \\
    & 2 &     & 0.95 & 0.93 & 0.94          & 0.88          & 0.69             & 0.66           \\
    \hline
    \end{tabular}
    \caption{%
        F1 scores of the modular memorization, LSTM and TED models, as well as of the end-to-end TED model. %
        For the results in the first row, "initial", we use the inferred intents without any simplifications or, in the end-to-end case, the plain text utterances. %
        For the results in rows 3 and 4, "simplified", we simplify the dialogue data, as explained in \S\ref{sec:multiwoz:ambiguous}. %
        The second column indicates the setting for \texttt{max\_history}. %
        Appendix~\ref{apx:scores} contains a more detailed table that also shows accuracy scores.%
    }
    \label{tab:results_multiwoz}
    \normalsize
\end{table*}
At this point, the memorization model achieves an F1 score of 0.84 on the training set, indicating that system actions are ambiguous (see Table~\ref{tab:results_multiwoz}). 
Throughout the remaining section we simplify the training data further, until all dialogues are consistent. 

The first problem leading to a consistency is that the availability of venues is not annotated, even though this knowledge is required to correctly choose the next action.
We add these annotations in the form of special ``status'' slots for each venue type, which take the values \verb+unique+, \verb+NA+, or \verb+available+, respectively for the three situations.
In addition, the status slots take the value \verb+booked+ after the system has booked a particular venue.
At this point, the representation of a typical MultiWOZ dialogue could look like the second example given in Appendix~\ref{apx:example}.
Adding status slots improves the memorization model's F1 score to 0.87. 

By examining the mistakes made by the model, we find that this low F1 score stems from the fact that multiple system actions are ``correct''.
For example, when the wizard has a few dining options available, she may
\begin{itemize}
    \itemsep0em 
    \item recommend one of the options,
    \item recommend one of the options and ask if there is anything else she can do,
    \item list all options,
    \item ask for more information from the user.
\end{itemize}
All of these actions are ``correct'' in the sense that they seem natural to the user, and which action is chosen depends on the mood and character of the person who takes the role of the system. 
Nevertheless, these actions are assigned distinct action types (\verb+Recommend+, \verb+Reqmore+, \verb+Select+, \verb+Request+), once again leading to ambiguous behaviour.
To remedy this issue, we recombine action labels as follows: 
The action types \verb+Inform+, \verb+Recommend+, \verb+Select+, and \verb+Request+ merge to the single type \verb+Reply+, and the action types \verb+Goodbye+, \verb+Welcome+, and \verb+Greet+ merge to the single type \verb+Welcome+.

Merging the action types in this way still does not remove all the unobservable information: The system actions are still ambiguous, as the memorization model's scores are only 0.90. 
Once more, we examine the mistakes that are made by the model and find that the action \verb+General-Reqmore+ is unpredictable.
Specifically, whether or not the system (wizard) asks if the user requires anything else is a random choice.  
Therefore, we remove all \verb+General-Reqmore+ actions from the dataset, unless it is the only action that the system takes in between user inputs.
This, again, increases the F1-score.

However, for the memorization model to reach an F1-score of 1.0, we have to get rid of all ambiguities in the dataset.
By identifying branch points in the tree of all dialogue histories and recursively removing ambiguous branches, we identify the largest subset of parsed MultiWOZ dialogues that is unambiguous.
This subset contains 1691 of the 8534 MultiWOZ dialogues (MultiWOZ 2.1 contains 10438 dialogues, but 1904 of those are not completely annotated and can therefore not be parsed).

We have now arrived at a dialogue dataset that is deterministic, i.e.\ the F1 score of the memorization model is 1.0, as can be seen in Table~\ref{tab:results_multiwoz}.
Note, that this resulting simplified dataset is not in any way more realistic than the original. 
The fact that it significantly differs from the original should only illustrate the severity of the issue of ambiguous system responses.

\subsection{History independence}
\label{sec:multiwoz:nohistory}

To establish the history independence of the MultiWOZ dialogues, 
in addition to the memorization model, we also train two other modular models: a long-short term memory (LSTM) model loosely following \cite{williamsHybridCodeNetworks2017} and the recently introduced Transformer Embedding Dialogue (TED) model~\cite{vlasovDialogueTransformers2019}.

For completeness, we also train the TED model in end-to-end (retrieval) mode, where it takes the history of plain text utterances as input, and picks an appropriate response from the list of all responses. 
To compute the scores of this end-to-end model, we associate the picked response with its action label(s).
While we could have used stricter evaluation metrics, such as human evaluation or the BLEU score \cite{papineni_bleu_2002}, this allows us to compare results directly to those of the modular approach. 
Again, our models are a means of investigating properties of the dataset and are not intended to improve upon the state of the art.

We observe that the end-to-end model performs consistently worse than the modular TED and LSTM models (see Table~\ref{tab:results_multiwoz}), which is not surprising since it has to solve the harder problem of mapping plain text utterances to plain text utterances, while the amount of training data is the same.

The history independence becomes apparent when we reduce the length of dialogue history that the three models take into account.
Specifically, reducing \verb+max_history+, as defined in \S\ref{sec:multiwoz:ambiguous}, from 10 to 2 barely changes the scores of either model, no matter if the dataset has been simplified (as described in \S\ref{sec:multiwoz:ambiguous}) or not (see Table~\ref{tab:results_multiwoz}).
Thus, none of the policies benefit from knowing the dialogue history beyond the last user input and its preceding system action.

Note, that \citet{vlasovDialogueTransformers2019} have shown that the TED model attends to  relevant pieces of a significantly longer dialogue history to predict the next system action. 
Thus, our result is indeed an issue of the dataset, not of the models used.

The apparent history independence is also plausible when we consider the excerpt from conversation \verb+MUL0104+, displayed in Appendix~\ref{apx:excerpt}.
Given the first two events (system and user utterance), anyone could predict the content of the subsequent system output. 
No further information would be required.
In particular, the remaining history of the conversation is irrelevant.
We observe the same situation repeatedly throughout the MultiWOZ dataset.
Furthermore, we also note that some of the best-performing models on MultiWOZ and similar datasets either neglect the dialogue history \cite{rastogiScalableMultidomainConversational2019, chaoBERTDSTScalableEndtoEnd2019}, or use a form of LSTM to encode it, which is naturally biased towards the most recent parts of the history \cite{mehriStructuredFusionNetworks2019}.

\section{Taskmaster-1}
\label{sec:taskmaster}

We repeat our analysis with Taskmaster-1, which by itself consists of two datasets: one which is collected via a Wizard-of-Oz setup, similar to MultiWOZ, and another for which each dialogue is written by a single human.
In this paper we only consider the latter, for which each dialogue concerns one of the following domains: Uber/Lyft ride bookings, movie ticket and restaurant reservations, coffee or pizza orders, and car repairs.
The Taskmaster-1 dialogues come with detailed annotations for utterance segments, including clues about the general intent and domain of the utterance in which they appear.

To remove the knowledge base dependence, we delexicalize annotated segments~\cite{mrksicNeuralBeliefTracker2016} and tag utterances with (i) the dialogue domain, (ii) the domain as classified by a simple regex, and (iii) the annotated segments that occur in the utterance. 
We then run the same end-to-end retrieval setup as we do with MultiWOZ.
Once more, we find that the resulting scores are almost history independent (see Table~\ref{tab:taskmaster} in Appendix~\ref{apx:scores}), which can also be concluded from reading the dialogues.
From analyzing prediction mistakes and reading the dialogues it is evident that the system responses are ambiguous in the same sense as in the MultiWOZ dataset.

\section{Conclusions}
\label{sec:conclusions}

We have analysed two recent dialogue datasets: MultiWOZ 2.1 and Taskmaster-1. 
Both datasets are purely human-generated and therefore contain natural utterances on both the user and system sides. 
In addition, both datasets contain useful annotations.
We show, however, that 
\begin{enumerate}
    \itemsep0em 
    \item both MultiWOZ and Taskmaster-1 are not suitable to train supervised dialogue systems on next-action prediction, unless they predict the \textit{probability distribution} of system actions instead of a single best action, and
    \item dialogues in these datasets are nearly history independent.
\end{enumerate}
The ambiguities in the action selection that make it impossible to train unique-response systems stem from the fact that the dialogues are highly dependent on the knowledge base, as well as on unobservable information such as the personality and mood of the wizard.
Furthermore, we hypothesize that the history independence stems from the greater problem that turkers are asked to pretend to want to achieve a goal.
Thus, they are not actually interested in the information they obtain, but are motivated only to complete each dialogue as soon as possible.

We suspect that instead of prescribing what the user ought to want, it would be better to describe a scenario to the user and let him/her explore the available options through interaction with the system. 
To remedy the ambiguities, an automatic system response should be enforced during data collection if the same dialogue state has been encountered before.

\subsubsection*{Acknowledgments}

We would like to thank the Rasa team and the Rasa community for feedback and support. 

\bibliography{multiwoz}

\appendix

\section{Appendices}
\label{sec:appendix}

\subsection{Example of a parsed conversation}
\label{apx:example}
The modular policies (memorization, LSTM, and modular TED) operate on training data as presented in the following example.
\small
\begin{itemize}
    \item \verb+inform{"hotel_area": "specific"}+
    \begin{itemize}
        \item \verb+Hotel-Select+
    \end{itemize}
    \item \verb+inform{"hotel_name": "specific"}+
    \begin{itemize}
        \item \verb+Hotel-Booking-Book{+ \verb+ "hotel_reference": "AHG32K"}+
        \item \verb+Hotel-Inform+
    \end{itemize}
    \item \verb+bye+
    \begin{itemize}
        \item \verb+General-Goodbye+
    \end{itemize}
\end{itemize}
\normalsize
Here, lines starting with a `\verb+*+' indicate a user turn and provide the user intent (\verb+inform+ or \verb+bye+) as well as the slots the user wishes to inform the system about.
The lines starting with a `\verb+-+' indicate a system action.

If status slots are provided, the example becomes
\small
\begin{itemize}
    \item \verb+inform{"hotel_area": "specific"}+
    \begin{itemize}
        \item \verb+Hotel-Select+
    \end{itemize}
    \item \verb+inform{"hotel_name": "specific"}+
    \begin{itemize}
        \item \verb+slot{"hotel_status": "unique"}+
        \item \verb+Hotel-Booking-Book{+ \verb+ "hotel_status": "booked",+ \verb+ "hotel_reference": "AHG32K"}+
        \item \verb+Hotel-Inform+
    \end{itemize}
    \item \verb+bye+
    \begin{itemize}
        \item \verb+General-Goodbye+
    \end{itemize}
\end{itemize}
\normalsize
where \verb+slot{...}+ denotes a slot being set by a knowledge base.

\subsection{Example excerpt}
\label{apx:excerpt}
This is an excerpt of conversation \verb+MUL0104+ from MultiWOZ 2.1:
\begin{itemize}
    \item ...
    \item There are two options - the University Arms Hotel in the centre and the Huntingdon Marriott Hotel in the west. Do you have a preference?
    \begin{itemize}
        \item The University Arms Hotel. Can you book that for 5 nights please?
    \end{itemize}
    \item What day would you like to stay and how many people will be staying?
\end{itemize}

\subsection{Detailed scores}
\label{apx:scores}

\small
\begin{table*}[tbp]
\centering
\begin{tabular}{|l|l|l|l|llll|}
\hline
Model & \verb+max_+ & $N$ & Note                  & \multicolumn{2}{l}{Training} & \multicolumn{2}{l|}{Testing} \\
       & \verb+history+ &            &                              & F1        & accuracy        & F1        & accuracy       \\
\hline\hline
\multirow{6}{*}{\rotatebox{90}{memorization}} & 10 & all & use inferred intents                 & 0.84 & 0.83 &           &                 \\
 & 10 & all & + add status slots                         & 0.87 & 0.87 &           &                 \\
 & 10 & all & + merge action labels                      & 0.90 & 0.90 &           &                 \\
 & 10 & all & + remove \verb+reqmore+                 & 0.91 & 0.91 &           &                 \\
 & 10 & 1691 &                   & 0.96 & 0.96 &           &                 \\
 & 10 & 1691 & + remove ambiguous dialogues                 & 1.00 & 1.00 & 0.45 & 0.63 \\
\hline
\multirow{8}{*}{\rotatebox{90}{TED}} & 10 & all & use inferred intents                 & 0.50 & 0.67 & 0.49 & 0.66 \\
 & 2 & all & & 0.47 & 0.66 & 0.45 & 0.65 \\
 & 10 & all & + add status slots       & 0.65 & 0.74 & 0.63 & 0.73  \\
 & 10 & all & + merge action labels   & 0.84 & 0.87 & 0.83 & 0.86 \\
 & 10 & all & + remove \verb+reqmore+   & 0.87 & 0.89 & 0.86 & 0.89 \\
 & 10 & 1691 &                  & 0.94 & 0.96 & 0.89 & 0.93 \\
    & 10 & 1691 & + remove ambiguous dialogues & 0.95 & 0.96 & 0.89 & 0.92 \\
    & 2 & 1691 &             & 0.94 & 0.94 & 0.88 & 0.91                \\
\hline
\multirow{8}{*}{\rotatebox{90}{LSTM}} & 10 & all & use inferred intents   & 0.51 & 0.67 & 0.51 & 0.67 \\
 & 2 & all &  & 0.47 & 0.66 & 0.47 & 0.66 \\
 & 10 & all & + add status slots  & 0.65 & 0.75 & 0.65 & 0.75 \\
 & 10 & all & + merge action labels  & 0.85 & 0.88 & 0.84 & 0.88 \\
 & 10 & all & + remove \verb+reqmore+  & 0.90 & 0.94 & 0.90 & 0.93 \\
 & 10 & 1691 &  & 0.93 & 0.96 & 0.90 & 0.94 \\
   & 10 & 1691 & + remove ambiguous dialogues & 0.95 & 0.97 & 0.92 & 0.95 \\
   & 2 & 1691 &        & 0.95 & 0.97 & 0.93 & 0.96  \\
\hline
\multirow{8}{*}{\rotatebox{90}{End-to-end TED}} & 10 & all & use plain-text utterances  & 0.22 & 0.61 & 0.21 & 0.61 \\
 & 2 & all &  & 0.22 & 0.62 & 0.21 & 0.61 \\
 & 10 & all & + add status slots & 0.28 & 0.64 & 0.28 & 0.64 \\
 & 10 & all & + merge action labels & 0.61 & 0.81 & 0.60 & 0.81 \\
 & 10 & all & + remove \verb+reqmore+ & 0.73 & 0.87 & 0.72 & 0.87 \\
 & 2 & all &  & 0.68 & 0.84 & 0.67 & 0.84 \\
 & 10 & 1691 &  & 0.68 & 0.82 & 0.64 & 0.80 \\
 & 10 & 1691 & + remove ambiguous dialogues & 0.69 & 0.84 & 0.66 & 0.82 \\
& 2 & 1691 & & 0.69 & 0.84 & 0.66 & 0.82 \\
\hline
\end{tabular}
\caption{%
Training and test scores on MultiWOZ 2.1 for the modular TED and LSTM models, as well as the TED-based end-to-end model and the memorization model.
The scores are presented for each model as the dataset is made increasingly consistent through various techniques described in \S\ref{sec:multiwoz:ambiguous} and denoted in the 'Note' column.
$N$ represents the number of dialogues used for training.%
}
\label{tab:all_results_multiwoz}
\end{table*}

\begin{table*}[tbp]
\centering
\begin{tabular}{|l|llll|}
\hline
\verb+max_history+ & \multicolumn{2}{l}{Training} & \multicolumn{2}{l|}{Testing} \\
             & F1       & accuracy & F1      & accuracy \\
\hline\hline
10           & 0.14 & 0.56 &  0.13  & 0.55 \\
2            & 0.12  & 0.54 & 0.10 & 0.53 \\
\hline
\end{tabular}
\caption{Training and test scores of the end-to-end TED model on 3000 training and 750 test dialogues from the Taskmaster-1 self-dialogues dataset. Results change little when \rm{max\_history} is reduced from 10 to 2.}
\label{tab:taskmaster}
\end{table*}
\normalsize


\end{document}